\documentclass[sigconf,authorversion,nonacm]{acmart}

\usepackage{graphicx,epsfig,subcaption,multirow,framed}
\usepackage{comment,breqn,setspace}
\usepackage{amsfonts,amsmath,amsthm}
\usepackage[linesnumbered,ruled]{algorithm2e}

\begin{document}

\title[On the Role of Multi-Objective Optimization to Transit Network Design]{On the Role of Multi-Objective Optimization\\ to the Transit Network Design Problem\vskip 0.1cm}

\author{Vasco Silva}
\email{vasco.dias.silva@tecnico.ulisboa.pt }
\affiliation{\institution{INESC-ID and IST, Univ. Lisboa}\country{}}

\author{Anna Finamore}
\email{annafinamore@gmail.com}
\affiliation{\institution{COPELABS, University Lusófona}\country{}}

\author{Rui Henriques\textbf{*}}
\email{rmch@tecnico.ulisboa}
\affiliation{\institution{INESC-ID and IST, Univ. Lisboa\newline \textbf{*}\textit{corresponding author}\vskip 0.1cm}\country{}}

\renewcommand{\shortauthors}{V. Silva, A. Finnamore, R. Henriques}

\begin{abstract}
Ongoing traffic changes, including those triggered by the COVID-19 pandemic, reveal the necessity to adapt our public transport systems to the ever-changing users' needs. 
This work shows that single and multi objective stances can be synergistically combined to better answer the transit network design problem (TNDP). Single objective formulations are dynamically inferred from the rating of networks in the approximated (multi-objective) Pareto Front, where a regression approach is used to infer the optimal weights of transfer needs, times, distances, coverage, and costs. As a guiding case study, the solution is applied to the multimodal public transport network in the city of Lisbon, Portugal. The system takes individual trip
data given by smartcard validations at CARRIS buses and METRO subway stations and 
uses them to estimate the origin-destination demand in the city. Then, Genetic Algorithms are used,
considering both single and multi objective approaches, to redesign the bus network that better fits the observed traffic demand. 
The proposed TNDP optimization proved to improve results, with reductions in objective functions of up to $28.3\%$. The system managed to extensively reduce the number of routes, and all passenger related objectives, including travel time and transfers per trip, significantly improve. Grounded on automated fare collection data, the system can incrementally redesign the bus network to dynamically handle ongoing changes to the city traffic.
\end{abstract}

\maketitle

\section{Introduction}
Mobility is a central aspect in modern societies, allowing people to take part in a
multitude of activities that are the identity of our days. Worldwide Public Transport Systems are placing efforts to meet emerging changes to urban traffic demand pushed by the ongoing COVID-19 pandemic, increasing population at urban centers, changing road infrastructures, among others. An efficient public transport network is, in this context, further essential to meet individuals' needs, promoting a sustainable urban mobility along its environmental, social and economical axes. 
In particular, this work addresses the Transit Network Design Problem (TNDP), where an efficient (multimodal) traffic network is one guaranteeing: i) comprehensive demand coverage, ii) low transfer rates and walking needs, iii) low travel times, iv) low waiting times, v) controlled operator costs, vi) low environmental footprint, and viii) safety guidelines such as maximum occupancy and recommended individual distances. 


State-of-the-art TNDP approaches, when faced with multiple objectives, generally resort to one of two major options. First, an enumeration of solutions at the Pareto front using a Multi Objective optimization stance, providing decision makers with multiple solutions \cite{jha2019multi,chen2006simulation,chen2010stochastic}. Despite their relevance, non-dominated solutions generally fail at providing an optimal compromise between all objectives, and existing works generally place strict assumptions on the variability of passenger demand. Second, complementary approaches pursue a Single Objective
Optimization process by placing an weighted sum of all the target objectives \cite{farahani2013a,bielli2002genetic,yu2005optimizing,jia2019urban,fan2006optimal}. In this context, weights are generally placed in accordance with the perceived importance of each objective, and further normalized according to their ranges of values. However, objective relevance is further dependent on additional confounding factors, including the extent of possible improvements per objective, turning this classic specification highly subjective. 

This work proposes explores why and how single and multi objective optimization approaches can be synergistically combined to addressed the enumerated challenges associated with the TNDP problem. 
In this work, weights are dynamically inferred from ratings placed on non-dominated network solutions uniformly sampled along the Pareto front approximation. The network ratings are thus used to replace subjective perceptions on the objectives' relevance, providing decision makers with better options to perform operational, tactic and strategic reforms in transport systems.

An end-to-end methodology for the TNDP task from raw automated fare collection (AFC) data gathered from multimodal public transport systems is further proposed, comprising principles for network processing, origin-destination matrix estimation, route generation, and efficient evaluation.  Genetic Algorithms, considered within both single and multi objective approaches, are outlined to redesign the transport network. 
Grounded on AFC data, the optimized network can be dynamically revised in the presence of more recent validations to handle ongoing changes to the city traffic. 

As a guiding case study, the proposed tool is applied to the Lisbon's public transport network in Portugal. The system takes individual trip data given by smartcard validations at CARRIS buses and trams, and METRO subway stations. 
The combined multiple-and-single optimization stance yields significant improvements in the design of the CARRIS bus network, with reductions in objective functions up to $28.3\%$. The gathered results show the possibility of using available fleet more efficiently to better serve passenger demand, moderately reducing transfer needs and travel times, while reducing the environmental impact of the network.

The manuscript is structured as follows. Section 2 provides essential background on optimization tasks. Section 3 surveys state-of-the-art contributions to the TNDP task. Section 4 describes the proposed end-to-end TNDP methodology. Section 5 discusses the results gathered from its application Lisbon's network. Finally, major concluding remarks are drawn.

\section{Background}

Finding the best elements, $\mathbf{x^{*}}$, from a set of alternatives, 
$\mathbb{X}$, according to a set of criteria, $F = \{f_{1}, f_{2}, ..., f_{m}\}$
is the essence of optimization. $\mathbf{x} = \{x_{1}, x_{2}, ..., x_{n}\}$ 
is a set of what is called design or decision variables.
$f_{i} : \mathbb{X} \mapsto Y, \quad (i=1, 2, ... m)$, with $Y \subseteq 
\mathbb{R}$, are the criteria or objective functions. These problems can
have constraints that limit the values that the design variables can take. 
We can formalize optimization problems as follows:

\vskip -0.2cm
\begin{equation}
\begin{split}
    \underset{\mathbf{x} \in \mathbb{X}}{\text{minimize}} \quad 
    f_{i}(\mathbf{x}) & , \quad (i=1, 2, ..., m)
    \\
    \text{subject to} \quad h_{j}(\mathbf{x})=0 & , \quad (j=1, 2, ..., o)     
    \\
    g_{k}(\mathbf{x})\leq0 & , \quad (k=1, 2, ..., p) \hspace{0.3cm}. 
\end{split}
\end{equation}

In Single Objective Problems ($m$=1), it is easier to distinguish the quality
of solutions. The lower the value of that
objective, the better the solution. However, when we have more than one objective, such distinction becomes harder to make. For example, if we have 
two solutions for a bi-objective problem, $\mathbf{x}^{1}$ and $\mathbf{x}^{2}$,
with $f_{1}(\mathbf{x^{1}}) < f_{1}(\mathbf{x^{2}})$ and
$f_{2}(\mathbf{x^{1}}) > f_{2}(\mathbf{x^{2}})$, we may not be able to identify the best solution without subsequent inspection. In this context, a multi objective problem 
can be converted into a single objective one by creating a new objective function, generally defined as:

\vskip -0.2cm
\begin{equation}
    f^{\prime} (\mathbf{x}) = \sum_{i=1}^{m} w_{i}f_{i}(\mathbf{x}) \hspace{0.2cm}.
\end{equation}\label{equation:multitosingle}
\vskip -0.1cm

This formulation, however, limits the variety of solutions we can find since
it does not reveal to the user where are the compromises between objectives. 
Pareto optimality becomes a relevant definition when dealing with 
multi objective problems as it provides us with a definition of optimality that
considers multiple objectives. A solution is Pareto optimal if we cannot improve 
one objective without damaging the quality of the remaining objectives. 
A Pareto optimal solution compromises the different objectives in an 
optimal way. Different Pareto optimal solutions represent different 
balances between the objectives. The set of all Pareto optimal solutions
is called the Pareto frontier. Pareto optimality is tightly coupled to 
the definition of domination. A solution $\mathbf{x^1}$ dominates another 
solution $\mathbf{x^2}$, in which case, we write $\mathbf{x^1} \prec 
\mathbf{x^2}$ if, and only if:

\vskip -0.2cm
\begin{equation}
\begin{split}
    \forall i \in \{0, ... , m\} : f_{i}(\mathbf{x^1}) \leq f_{i}(\mathbf{x^2}) &
    \quad \wedge
    \\
    \exists j \in \{0, ... , m\} : f_{j}(\mathbf{x^1}) < f_{j}(\mathbf{x^2}) & .
\end{split}
\end{equation}

Now we can define the Pareto frontier, $\mathbf{X^*}$ as the set of points that are not dominated by any other solution. In other words:

\vskip -0.2cm
\begin{equation}
    \mathbf{x^*} \in \mathbf{X^{*}} \Leftrightarrow \nexists \mathbf{x}
    \in \mathbb{X} : \mathbf{x} \prec \mathbf{x^*}\hspace{0.1cm}.
\end{equation}

While Single Objective approaches try to find the best solution according to a single
criteria, Multi Objective approaches approximate the Pareto front to identify compromises between objectives
and assess their quality.

Genetic Algorithms (GA) are population-based optimization algorithms with well-established success towards both Single and Multi Objective ends. GA 
mimic nature's evolutionary process. Solutions are individuals in a population
represented as a string of symbols that encode a solution for the problem
we are trying to solve. Individuals reproduce and mutate to give place to new
individuals. Reproducing means combining the genetic code of two individuals to
make new solutions that are based on their parents and mutation means randomly
changing the genetic code of an individual to introduce variety and try to escape
local minimums. As a result, the quality of populations is expected to increase along iterations.  The Classic Genetic Algorithm was one of the first to use these principles in single objective optimization by John H. Holland
in 1976 \cite{geneticalgorithms}. 
The Non-dominated 
Sorting Genetic Algorithm II is a multi objective optimization algorithm 
proposed by Deb et al. \cite{deb2002fast}. These are the algorithms extended
in the present work.
\vskip -0.5cm

\section{Related Work}

Newell \cite{newell1979some} points out the non-convex properties of the Transit Network Design Problem (TNDP),
showing for instance that changing bus frequency (operation costs) may not impact user costs. Baaj and
Mahamassi  \cite{baaj1991ai} further highlights the hard combinatorial nature of the
problem caused by its discrete nature. In contrast with bus frequencies, lightly depicted as numeric decision variables in problem
formulations, representation of network routes is more challenging. This makes Mixed Integer
formulations, like the one proposed by Wan and Hong \cite{wan2003mixed}, rather ineffective to solve TNDP. In Wan and Hong's formulation,
the presence of each node in each route is modeled as a binary variable. This
makes it so that a small network with 10 nodes and 19 links, ends up having 
363 binary decision variables, 30 integer decision variables and 303 continuous
decision variables when solving TNDP. In this context, metaheuristic approaches are necessary. 
While some works try to change the configuration of routes
in a bus network. Others try to optimize both the route configuration and 
their respective working frequencies simultaneously, the so referred 
Transit
Network Design and Frequencies Setting Problem (TNDFSP). 
A significant number of works have taken different approaches to solve these problems. 

Most works reduce the problem to a single
objective formulation (Eq.\ref{equation:multitosingle}) and then
solve the single objective variant \cite{farahani2013a}, reducing the variety
of solutions found by these works. 
Yu et al. \cite{yu2005optimizing} used Ant Colony Optimization to TNDP, maximizing passenger flow and minimizing transfers. In their model, OD pairs work as the nest and the 
food source for several sub-colonies and the pheromone trail is based on 
passenger density. They tested their work on an existing network
in the city of Dalian (3,200 links and 2,300 nodes with 89 
lines), showing the possibility to reduce the number of lines to 61 without
compromising satisfied demand. 
Pattnaik et al. \cite{pattnaik1998urban} used GA to solve TNDFSP, minimizing traveling times and operating time of buses, given feasibility constraints. Their approach worked
in two phases, first a Candidate Route Set Generation Algorithm (CRGA) followed by the GA identification of optimal routes and frequencies based on Fixed and Variable String Length codifications of routes. Bielli et al. \cite{bielli2002genetic} used GA to solve TNDFSP using considerably different modeling principles. 
Bielli et al. use a fixed length representation in which each chromosome is
a network and each gene represents a pre-generated line with a frequency 
and an on/off switch that indicates whether the route is active in the 
network. This means that every pre-generated route will exist in every
network but with different frequencies and it can be active or not.
Their formulation places a weighted sum of efficacy, efficiency
and quality of service metrics. The work shows significant improvements Parma city, 
Italy. 
Chien et al. \cite{chien2001genetic} also used GA in the context of TNDFSP
but with a different goal: 
optimizing feeder bus routes. Feeder routes bring people from (to) several points
to (from) a central hub. Their objective function also minimizes user costs (access, waiting and in-vehicle costs) and
operator costs (round trip time of
the buses and the headway). 
They tested their approach in small
networks and found that, when the parameters are properly tuned, the GA
can find the optimal route. 
Fan and Machemehl \cite{fan2006optimal} also used GA to solve TNDFSP. 
Their objective function considers, user costs, operator costs and 
unsatisfied total demand costs, with the relative importance easily parameterized. 
Similarly to Pattnaik et al., 
their approach uses Dijkstra's algorithm
and Yen's k-shortest paths algorithm to generate routes between OD pairs. 
The authors show the role of GA methods against alternative population-based methods. 
Jia et al. \cite{jia2019urban} tackles TNDP for the bus network in Xi'an, a 
city in Northwest China, showing the role of Complex Network theory to improve sustainability and robustness without directly modeling demand. 

In contrast, a scarcer number of TNDP approaches place multi-objective optimization stances in order to find non-dominated solutions of potential interest, providing stakeholders with multiple options. 
Jha et al. \cite{jha2019multi} approximated the TNDFP Pareto front combining average passenger time and operating costs using Intelligent Water Drops (IWD) algorithm, a novel and tailored evolutionary optimization technique. Chen et al. \cite{chen2010stochastic} proposed a stochastic framing for multi-objective optimization for designing transportation networks under demand uncertainty. Arbex et al. \cite{arbex2015efficient} introduced an Alternating Objective Genetic Algorithm (AOGA) to boost efficiency, in which the objectives are cyclically alternated along the generations. 
Wang et al. \cite{wang2020multi} formulates the TNDP as a multi-level task described by the skeleton network, arterial network, and feeder network, proposing an hybrid heuristic approach. 
Despite the relevance of multi-optimization stances to assess alternative solutions, existing works generally focus on road developments and reforms \cite{brauers2008multi}, including street constructions, lane additions, allocations and reorientations \cite{miandoabchi2012hybrid}. Zhang et al. \cite{zhang2020multi} seek non-dominated solutions for railway alignment considering costs and environmental impacts; Inti et al. pursued \cite{inti2021sustainable} non-dominated road developments balancing environmental, economic and constructability factors; and Sohn \cite{sohn2011multi} identified road diet network solutions maximizing utility for cyclists with minimum impact on motorists.

Despite the relevance of the surveyed principles, Single Objective stances depend on the subjective and difficult task of weighting conflicting objectives; while Multi Objective stances generally fail at providing an adequate compromise between all objectives. In addition, most works focus on single transport modes and place strict demand assumptions, not assessing the impact of origin-destination demand variability, possibly given by all raw passenger trips in a (multimodal) transport system. 

\section{Solution and Problem Formulation}

\begin{figure*}
    \centering
    \includegraphics[width=0.67\linewidth]{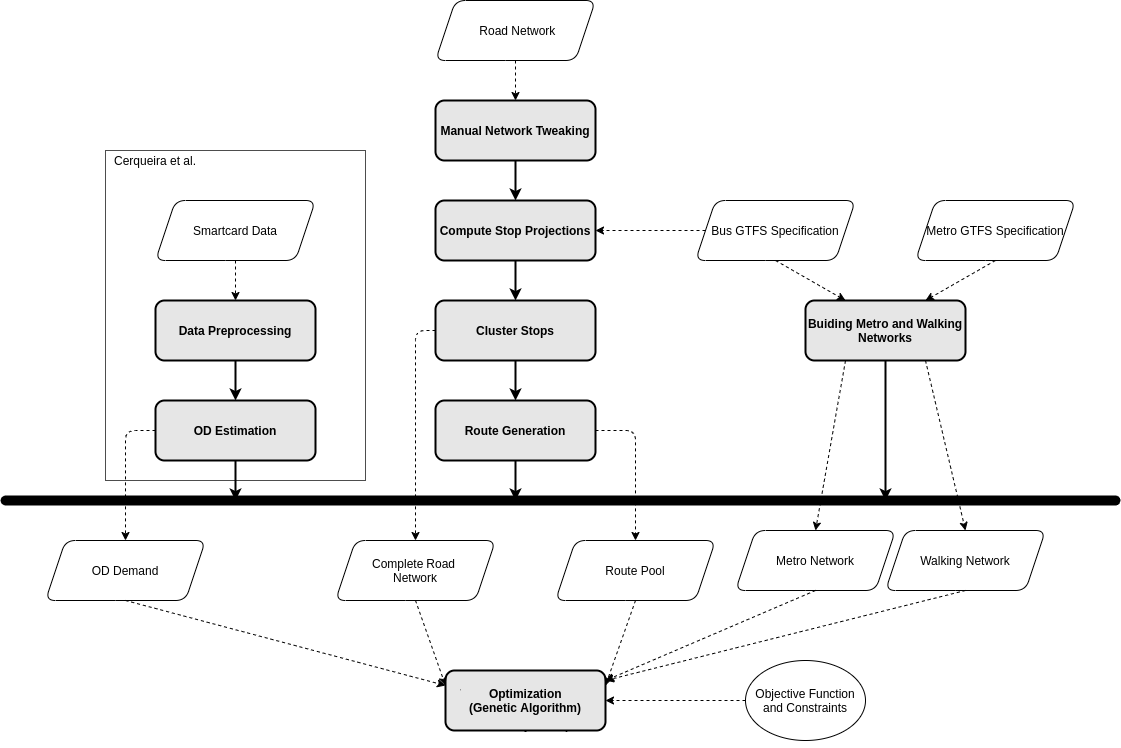}
    \caption{Proposed route optimization methodology.}
    \label{fig:final_solution}
\end{figure*}

This section introduces an end-to-end methodology to answer the targeted transit network design problem (TNDP), covering the essential aspects pertaining to: the inference of traffic demand from smart card validation; the consolidation of the road network with the public transport networks from different operators; the definition of evolutionary operators and principles for the effective generation of routes; the assignment of origin-destination needs to transportation means; and the formulation of the targeted multi and single objective approaches. An overview of the developed solution can be seen in Figure \ref{fig:final_solution}.

\subsection{Preprocessing}

To run an optimization process on a bus network that operates on top of a 
road network, we first have to have the available bus stops as nodes on
a the road network. That way, we can then define a bus route as a sequence
of stops and assume that the buses travel between the stops through the 
optimal time path. The road network was retrieved from the Open Street
Map (OSM) database through the \textsf{OSMnx} python package which gives
us the network as a \textsf{NetworkX} graph. The stop locations were 
taken from the CARRIS General Transit Feed Specification (GTFS). Since
the stops and networks are built on inaccurate GPS readings, fixing the
stops on the network is a complex task. For that end, we used the work
of Vuurstaek et al. \cite{vuurstaek2018gtfs}.

There are stops that exist in close proximity to others and whose
only purpose is to provide people with multiple boarding points
so that stops do not become overcrowded. These usually serve 
different routes, but from an optimization point of view, they
are equivalent because they provide entry and alight points
in the same general areas. Because we have routes that are 
generated randomly, we want to have all these stops clustered
under the same stop so we do not generate redundant routes.
To that end, two stops, $u$ and $v$ are clustered into one
if the edge $(u, v)$ exists in the road network and has
a cost of less than $100m$ and if $deg_{in}(v)=1$.
We force the edge $(u, v)$ to exist because, if a road node
exists between the two stops, then that means that a route 
can go from $u$ to one other stop, $w$, without necessarily going
through $v$, making it so that, if stops $u$ and $v$ are
clustered under the location of $v$, that route would be 
way longer than it should as it would go from $v$ to $w$.
We force $deg_{in}(v)=1$ because if there is a route that
goes from a third stop $w$ to $v$ and we cluster $v$ and
$u$ under the location of $u$ then that route would have
to go from $w$ to $u$, potentially making it much longer
than what it, in fact, is. The $100m$ requirement is 
completely adjustable and it is there only so we do not
cluster stops on different ends of the same avenue.
If stops $u$ and $v$ can be joined and stop $v$ can be 
joined with a third stop $w$, then all the stops can
be joined under the same cluster. 
We started with $2193$ stops and managed to reduce this
number to $1786$ using this criteria to join stops.

The origin-destination estimation is also an essential
step since buses require smartcard validation only upon
entering. In this work, the estimators of Cerqueira et al. 
\cite{cerqueira2020estimators} were applied to produce OD
matrices which then serve as input to our optimization processes.

\subsection{Relevant networks and interactions}

In this work, we use several different graphs to represent everything we need
from the domain. It is then important to understand the different types of 
graphs, what is their purpose, and how they interact.

The road network, $G_{r}$, has edges representing road segments and nodes 
representing road junctions, road ends or bus stops.

The bus network, $G_{b}$, is the only one with several instances. 
This is the object of optimization and, as such, a genetic 
algorithm population is filled with networks of this type, i.e.
at each time step $t$, we will have a population $P_{t} = \{ G^{1}_{b}, 
G^{2}_{b}, ..., G^{n}_{b}\}$. These networks are built from a set of 
routes. A route is just a sequence of stops. All the tram routes are
included in the bus networks but they are never interfered with during
the optimization process. The edges in this network are identified 
by an origin stop, a destiny stop and a route making the connection.

The metro network, $G_{m}$, also has edges identified by an origin station,
a destiny station and a line color connecting the two. We want to assess 
what the use of the whole transportation network would be like with the bus 
network under evaluation, so the metro network must be present. It 
will not be changed and it is a singleton network. The metro trains
are assumed to travel at $60Km/h$.

Finally, the walking network, $G_{w}$, connects the bus network with the metro 
network and close bus stops. It is assumed that people are willing to 
walk up to $300m$ to transfer. The nodes are bus stops or metro stations
and the edges represent possible walks between them. The walks are 
assumed to be a straight line from the origin to the destiny. People 
are assumed to walk at $5Km/h$. This network is never changed.

During optimization, we integrate the bus network, under evaluation,
with the walking and metro networks, creating
a complete multimodal network, $G_{c}$, in which trips that involve walking, 
bus and metro can be planned.

\subsection{Network evaluation}\label{sec:networkevaluation}

In order to assess transfer needs, travel times, waiting times and all 
the other objectives we are striving to improve, we need to know how the
public would use the transportation network under evaluation. In this work,
passengers are grouped according to their origin-destination (OD) pair. 
To reduce computational complexity, OD units are not individual bus stops
or metro stations, instead Lisbon is divided into a $30\times30$ grid
and the squares are the OD units. Everyone moving between the same
pair of squares is assumed to do so through the same trip. 
The trips are a result of a computation of time optimal
paths in the complete network, $G_{c}$, between all OD pairs.
When a bus network is able to connect an origin and a destination, it does
so through a trip, which is a path in the complete network, $G_{c}$. For
further analysis of the trips, we can divide the trip into stages. There are
three different kinds of stages, a bus stage, a walking stage and a metro stage.
Practically speaking, a bus stage is a path, along a single route, in the bus
network, a metro stage is a path, along a single line, in the metro network and
a walking stage is a single edge path that connects two bus stages or a metro stage
and a bus stage. This distinction is important because, every time there is 
a stage change, we apply a transfer penalty and the path cost increases. This 
penalty intends to simulate user's preference for trips with less transfers if
that doesn't incur a big increase in the overall trip time.

Now, we establish important notation for sequent formulations: 

\begin{itemize}
\setlength\itemsep{-0.01cm}
  \item $W^{O}$ - the origin set of geographies;
  \item $W^{D}$ - the destiny set of geographies;
  \item $W$ - set of all origin-destination pairs;
  \item $q(s, t)$ - amount of passengers traveling between $s$ and $t$;
  \item $R$ - set of all routes in a bus network;
  \item $\mathcal{T}$ - set of all trips between all OD pairs,\newline
  $\mathcal{T}$=$\big \{ T(s, t) \mid s \in W^{O}, t \in W^{D}\big\}$;
  \item $T(s,t)$ - trip from $s$ to $t$, a sequence of stages,\newline
  $T(s, t) = (T^{0}(s, t), \dots, T^{n}(s, t)),$  where $n \in \mathbb{N}_{0}^{+}$;
  \item $T^{k}(s, t)$ - the $k^{th}$ stage on the trip from $s$ to $t$. 
  It is defined as 
  sequence of triplets of the form $(u, v, r)$ where $u$ and $v$ are adjacent 
  stops or stations and $r$ is the route/line connecting them or a special
  marker indicating that the path from $u$ to $v$ was made by foot;
  \item $T_{bus}(s, t)$ - bus stages in the trip from $s$ to $t$;
  \item $T_{metro}(s, t)$ - metro stages in the trip from $s$ to $t$;
  \item $T_{walk}(s, t)$ - walking stages in the trip from $s$ to $t$;
  \item $t(s, t)$ - travel time between nodes $s$ and $t$,\newline i.e.
  $\ t(s, t) = t_{inv}(s, t) + t_{wai}(s, t) + t_{wal}(s, t)$;
  \item $t_{inv}(s, t)$ - in-vehicle time between nodes $s$ and $t$;
  \item $t_{wai}(s, t)$ - waiting time between nodes $s$ and $t$;
  \item $t_{wal}(s, t)$ - walking time between nodes $s$ and $t$;
  \item $f_{r}$ - frequency of service in route $r$ (in buses/hour);
  \item $t_{r}$ - time it takes for a bus to go from the starting station
  to the terminal station in a given route;
  \item $l_{r}$ - length of route $r$;
  \item $h$ - the number of hours the network is active per day.
\end{itemize}
\vskip 0.1cm

Now we can define some quality metrics regarding a bus network.
The total length, TL, of a bus network is computed as:

\vskip -0.15cm
\begin{equation}
    TL(G_{b}) = \sum_{r \in R}{l_{r}}\hspace{0.2cm}.
\end{equation}
\vskip 0.1cm

The Unsatisfied Demand, UD, of a bus network is defined as:

\vskip -0.1cm
\begin{equation}
    UD(G_{b}) = 1 - \frac
    {\sum_{(s, t) \in W} {q(s, t) \times \text{CO}(G_{b}, s, t)}}
    {\sum_{(s, t)\in W} {q(s, t)}} \hspace{0.2cm},
\end{equation}
\vskip 0.1cm

\noindent where $CO(G_{b}, s, t)$, the cover function, is one if the bus network 
$G_{b}$ provides, along with the rest of the transportation network, 
a connection between the origin $s$ and destiny $t$ within a number 
of transfers bellow the maximum allowed and zero otherwise.

In vehicle time (IVT) is computed as follows:

\vskip -0.15cm
\begin{equation}
    IVT(G_{b}) =  \sum_{(s, t) \in W} t_{inv}(s, t) \cdot q(s, t)
    \hspace{0.2cm}.
\end{equation}
\vskip 0.1cm

The Average Number of Transfers per passenger, ANT, is computed as follows:

\vskip -0.15cm
\begin{equation}
    ANT(G_{b}) = \frac
    {\sum_{(s, t) \in W} \big( |T(s, t)| - |T_{walk}(s, t)| - 1 
    \big) q(s, t)}{\sum_{(s, t) \in W} q(s, t)}\hspace{0.1cm}.
\end{equation}
\vskip 0.1cm

\subsection{Problem formulation and modeling}

Now, we can define the optimization problems to solve. The topology
optimization or the Transit Network Design Problem (TNDP) is modeled as:

\vskip -0.15cm
\begin{equation}
\begin{split}
    \underset{G_{b}}{\text{minimize}} \quad & 
    TL(G_{b}), \quad UD(G_{b}), \quad IVT(G_{b}), \quad ANT(G_{b}) , 
    \\
    \text{subject to} \quad & l_{r} \leq \text{MAX\_ROUTE\_LEN} \quad \forall r \in R ,    
    \\
    \quad & l_{r} \geq \text{MIN\_ROUTE\_LEN} \quad \forall r \in R,
    \\
    \quad & |R| \leq \text{MAX\_NUMBER\_ROUTES},
    \\
    \quad & |R| \geq \text{MIN\_NUMBER\_ROUTES}.
\end{split}
\end{equation}

In terms of genetic modeling, our proposal uses principles similar to 
the ones proposed by Pattnaik et al. \cite{pattnaik1998urban} VSLC.
We have a route pool which has the original
routes found in the CARRIS network and a set of generated routes. Every 
network has all the tram routes (fixed routes) that are in the original CARRIS network.

The initial population is an array of randomly sized
networks with routes randomly selected from the route pool. We never
commit to a predefined size. Route insertion and deletion operators
are introduced in the mutation process. Mutating consists of swapping
a random route in the network with a random route in the route pool.

There are two types of generated routes, hub connectors and traversal routes.
Hub connectors are routes connecting the busiest stops through the shortest
paths and traversal routes are longer routes whose purpose is to enable 
easier connections between opposite sides of the city.

\subsection{Single objective formulations}

After we assess the quality of the solutions that are given by the multi
objective formulations, we can have a sense of what we are looking for
in a network. After we know that, we can try to get as close as possible
to a global optimum that represents the compromise we are looking for in
a network. To this end, we will be rating networks generated by NSGA-II.
The rating will then serve as reference for a linear regression that will
allow us to infer the weights for a weighted single objective formulation 
(Equation \ref{equation:multitosingle}) that stand for what we are 
looking for. We are trying to estimate:

\vskip -0.25cm
\begin{equation}
    \mathbf{w} = (w_{0}, w_{1}, w_{2}, \dots, w_{m})^{T},
\end{equation}

\noindent so that we can have a finely tuned objective function, aligned
with the perceived needs:

\vskip -0.25cm
\begin{equation}
    f(G_{b}) = w_{0} + \sum_{i=1}^{m} w_{i}f_{i}(G_{b}) \quad.
\end{equation}

To this end, the find the best weight vector $\mathbf{w}$
such that a given error function, $E(\mathbf{w})$, is minimized. 
The error functions measure the difference between the target and estimated values, 

\vskip -0.2cm
\begin{equation}
    y(\mathbf{x}) = \mathbf{w}^{T} \cdot \mathbf{x} \hspace{0.2cm}.  
\end{equation}

Minimizing the error is also an optimization problem, but in
this case, $\nabla E(\mathbf{w}) = 0$ is solvable when considering a squared error function, with optimal weights given by:

\vskip -0.2cm
\begin{equation}
    \mathbf{w} = (X^{T} \cdot X)^{-1} \cdot X^{T} \cdot \mathbf{t}\hspace{0.2cm},
\end{equation}

\noindent where $X$ is the design matrix. 
Assuming that we have $n$ bus networks from the NSGA-II algorithm, we
have, for our particular TNDP problem:

\vskip -0.2cm
\begin{equation}
    X = 
    {\scriptscriptstyle
    \begin{bmatrix} 
    1 & \text{\small TL}(G^{(1)}_{b}) & \text{\small UD}(G^{(1)}_{b}) & \text{\small IVT}(G^{(1)}_{b}) & \text{\small ANT}(G^{(1)}_{b})
    \\[0.3cm]
    1 & \text{\small TL}(G^{(2)}_{b}) & \text{\small UD}(G^{(2)}_{b}) & \text{\small IVT}(G^{(2)}_{b}) & \text{\small ANT}(G^{(2)}_{b}) 
    \\
    \vdots & \vdots & \vdots & \vdots & \vdots 
    \\
    1 & \text{\small TL}(G^{(n)}_{b}) & \text{\small UD}(G^{(n)}_{b}) & \text{\small IVT}(G^{(n)}_{b}) & \text{\small ANT}(G^{(n)}_{b}) 
    \\
    \end{bmatrix} 
    }
\end{equation}

\noindent and
\vskip -0.6cm 

\begin{equation}
    \mathbf{t} = \Big(
    max-r\big(G^{(1)}_{b}\big), 
    max-r\big(G^{(2)}_{b}\big), \dots , 
    max-r\big(G^{(n)}_{b}\big)
    \Big)^{T},
\end{equation}

\noindent where $r\big(G^{(i)}_{b}\big)$ is the average rating given to
the $i^{th}$ network and $max$ is the maximum rating a network can be given. 
The target for each network is $max$ minus its rating because we want the 
objective function to be minimized and so, the better the rating, the lower
the objective function value should be.

\section{Results}

\begin{figure}[!h]
    \centering
    \includegraphics[width=1\linewidth]{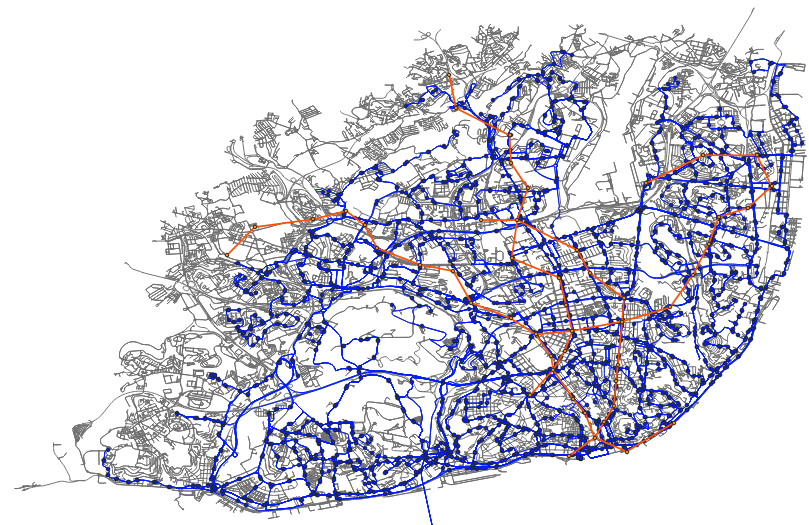}
    \caption{CARRIS network (blue) in October 2019 over the road network
    and Lisbon METRO network (orange).}
    \label{fig:lisbonnetwork}
\end{figure}

In this work, we work with traffic data from October 2019. 
The CARRIS bus network deployed at that time can be seen in
Figure \ref{fig:lisbonnetwork}. The network has 309 routes 
and 2,193 stops. Each route has, on average, 26.2 stops and 
each station, on average serves 3.7 routes. For the month of
October 2019, we have 6.2 million smartcard validations at
the bus entrances, spanning 4 days. These validations, as 
well as validations on the Metro network, are used as the
input for the work of Cerqueira et al. \cite{cerqueira2020estimators}
for inferring the OD demand.

The results presented in this section were obtained with genetic algorithm experiments
ran during $300$ iterations, with a population size of $200$, a mutation probability
of $0.1$ and, when applicable, a crossover probability of $0.8$. These parameters were
decided by doing sensibility analysis on a smaller data sample.

\begin{figure}[t!]
    \centering
    \includegraphics[width=\linewidth]{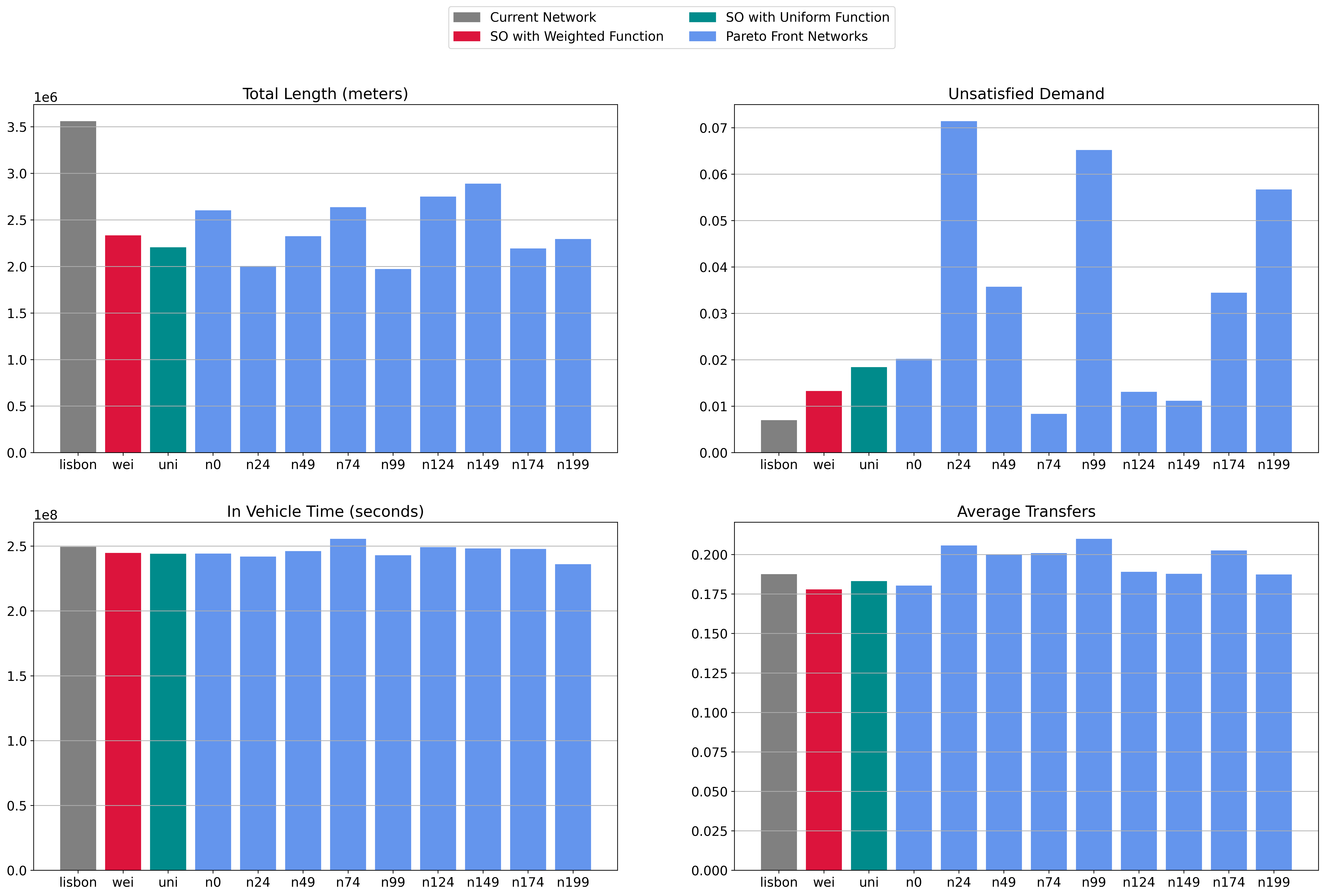}
    \caption{
    \small
    Comparing individual objectives of networks from Multi Objective 
    Optimization and Single Objective Optimization.
    "lisbon" is the original bus network. "wei" was obtained in a single 
    objective optimization considering the weighted function. "uni" was obtained 
    in a single objective optimization considering the uniform function. The 
    remaining networks resulted from NSGA-II.}
    \label{fig:objective_histograms}
\end{figure}

The networks that we discuss in this section were obtained with
the NSGA-II algorithm. The maximum allowable 
number of routes is 400 and the minimum is 200. A sample of 9 networks
was chosen for a careful examination and their objective function values
can be seen side by side in Figure \ref{fig:objective_histograms}. These
were subjected to a rating  on a scale of 1 to 10 so that we can infer the weights
for a single objective optimization process. The networks are identified 
by their order in the crowd distance sorting. The number of routes in each
network can be seen in Table \ref{table:numberroutes}.

\begin{table}[!h]
\footnotesize
    \begin{center}
    \begin{tabular}{r |*{5}{c}} 
        \hline
        network & lisbon & $n0$ & $n24$ & $n49$ & $n74$\\
        \hline
        route count & $309$ & $201$ & $200$ & $200$ & $209$ \\ 
        \hline \hline
        network & $n99$ & $n124$ & $n149$ & $n174$ & $n199$ \\
        \hline
        route count & $200$ & $207$ & $232$ & $200$ & $200$ \\
        \hline
    \end{tabular}
\end{center}
\caption{\label{table:numberroutes}\small Number of routes per network using the NSGA-II algorithm.
}
\end{table} 
\vskip -0.5cm

The big majority of the networks has a number of routes very close to the 
minimum allowable number of routes. The network with the maximum number 
of routes is  composed of 256 routes. We were expecting a more diverse 
set of networks when it comes to the number of routes, however the number
of routes is not an objective, instead we opted to use the total length 
of the network as a measure of route efficiency so the diversity lies in
the average length of the routes and not in the number  of routes. The 
average route length varies from $9.5Km$ to $13.3Km$ and presents a 
bimodal distribution with peaks around the $10.5Km$ and $12.5Km$ marks.

All the networks present a total length bellow the original network but no
network is capable of satisfying demand better. In terms of in-vehicle time
and average transfers, there are networks in the Pareto Front capable of 
more direct and shorter trips but there are also networks that demand longer
trips and a higher transfer rate.

Four experts were asked to rate the networks selected from 1 to 10. The average ratings for each network are depicted in Table \ref{table:ratings}.
\vskip -0.1cm
\begin{table}[h!]
\footnotesize
    \begin{center}
    \begin{tabular}{r |*{5}{c}} 
        \hline
        network & $n0$ & $n24$ & $n49$ & $n74$ & $n99$ \\
        \hline
        rating & $7.5$ & $3.5$ & $5$ & $5.75$ & $3.25$ \\ 
        \hline \hline
        network & $n124$ & $n149$ & $n174$ & $n199$ & \\
        \hline
        rating & $6$ & $7.5$ & $5.5$ & $4.5$ & \\
        \hline
    \end{tabular}
\end{center}
\caption{
\label{table:ratings}\small
Average rating of each network obtained with the NSGA-II algorithm.
}
\end{table} 
\vskip -0.5cm

After the linear regression we get the following weights:
\vskip -0.1cm

\begin{table}[h!]
\footnotesize
    \begin{center}
    \begin{tabular}{*{5}{c}} 
        \hline
         $w0$ & $w1$ & $w2$ & $w3$ & $w4$ \\
        \hline
         $-15.85$ & $1.04\times10^{-06}$ & $53.13$ & $7.75\times10^{-09}$ & $72.38$ \\
        \hline
    \end{tabular}
\end{center}
\caption{
\label{table:linearweights}\small 
Objective function weights after Linear Regression.
}
\end{table} 
\vskip -0.5cm

We ran a Classic GA with the weights in Table \ref{table:linearweights}. 
We ran the Classic GA again, but with the individual 
objectives summed and normalized with a min-max normalization so we could see the differences
between the quality of the solutions when we emphasize different objectives 
against uniform weights. The objective function in this case becomes:
\vskip -0.1cm
\begin{equation}
    f(G_{b}) = \sum^{m}_{i=1} \frac{f_{i}(G_{b}) - \min{(f_{i})}}
    {\max{(f_{i})}- \min{(f_{i})}} \hspace{0.1cm}. 
\end{equation} \label{equation:singleobjectivenormalize}

\begin{figure}[!t]
    \centering
    \includegraphics[width=.85\linewidth]{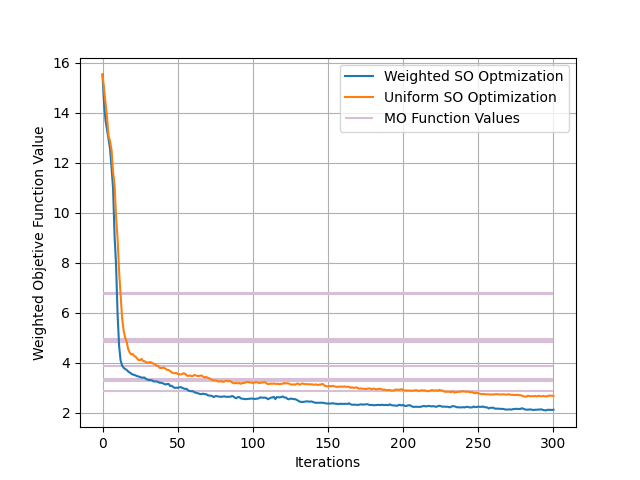}
    \vskip -0.1cm
    \caption{\small Average objective value along the iterations for the
    Single Objective GA considering Weighted function in blue and the Uniform
    function in orange. }
    \vskip -0.1cm
    \label{fig:singleobjectiveiterations}
\end{figure}

The average weighted objective function value over the algorithm iterations 
can be seen in Figure \ref{fig:singleobjectiveiterations} for both
experiments ran.
According to the criteria we set when rating the networks, when we use the 
weighted objective function, a Single Objective GA is able to produce networks
that are better than the best network we got from the Multi Objective
Optimization after about 60 iterations. When we use a uniform weight distribution
with normalized objectives we can reach a level of quality superior to the 
Multi Objective Optimization, yet only after about 200 iterations.

\begin{figure}[!t]
    \centering
    \includegraphics[width=.85\linewidth]{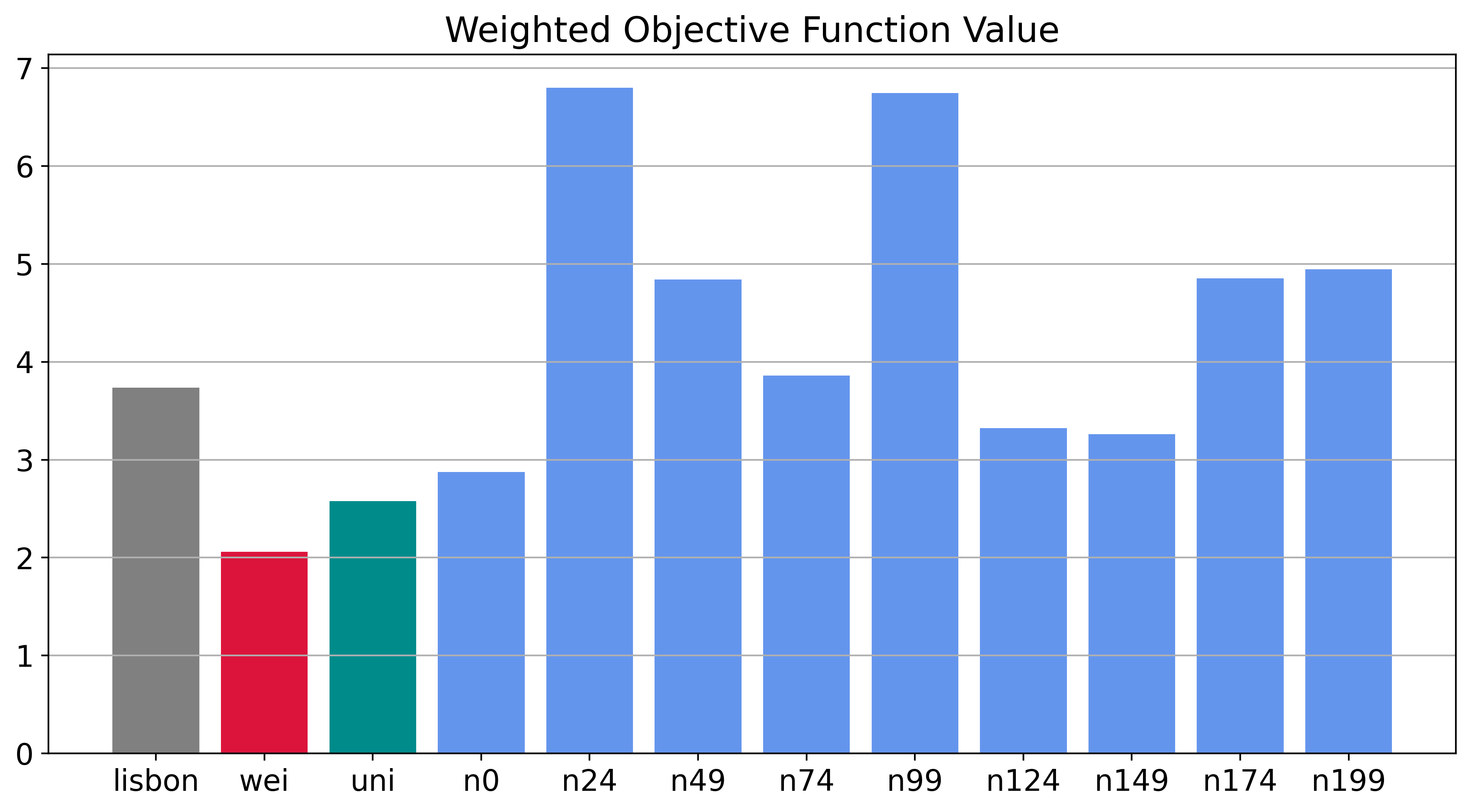}
    \vskip -0.1cm
    \caption{\small Weighted Objective Function Value for the original network and 
    the networks obtained through optimization.}
    \label{fig:individualobjectivecomparison}
    \vskip -0.3cm
\end{figure}

In Figure \ref{fig:individualobjectivecomparison} we can see the individual
objectives of the networks obtained through Single Objective Optimization 
side by side with the ones selected from the Multi Objective Optimization 
and the original CARRIS network. Both networks obtained with Single Objective
Optimization have $200$ routes. The highest rated network was network $0$
and, as we can see, the network obtained with the weights from the linear
regression trumps that network in every objective. The network obtained
with uniform weights and normalized objectives does not achieve the same 
success in the number of transfers but it manages to have smaller total 
length than the network obtained with a weighted function.

\begin{figure}[!t]
    \centering
    \includegraphics[width=.8\linewidth]{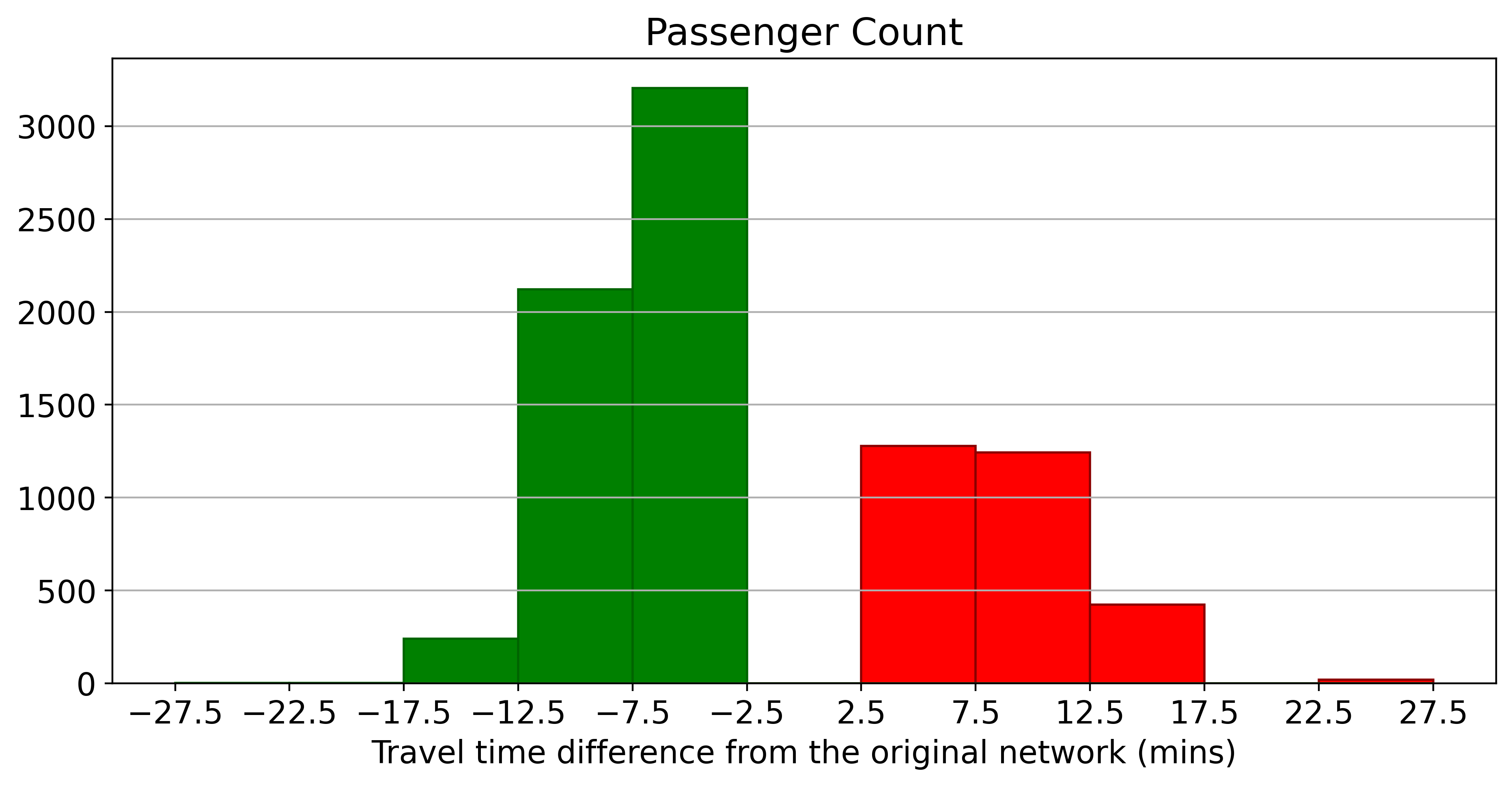}
    \vskip -0.1cm
    \caption{\small Differences \textit{in travel time} from the original network to the best network.}
    \label{fig:traveltimedifference}
    \vskip -0.3cm
\end{figure}

\begin{figure}[!t]
    \centering
    \includegraphics[scale=0.31]{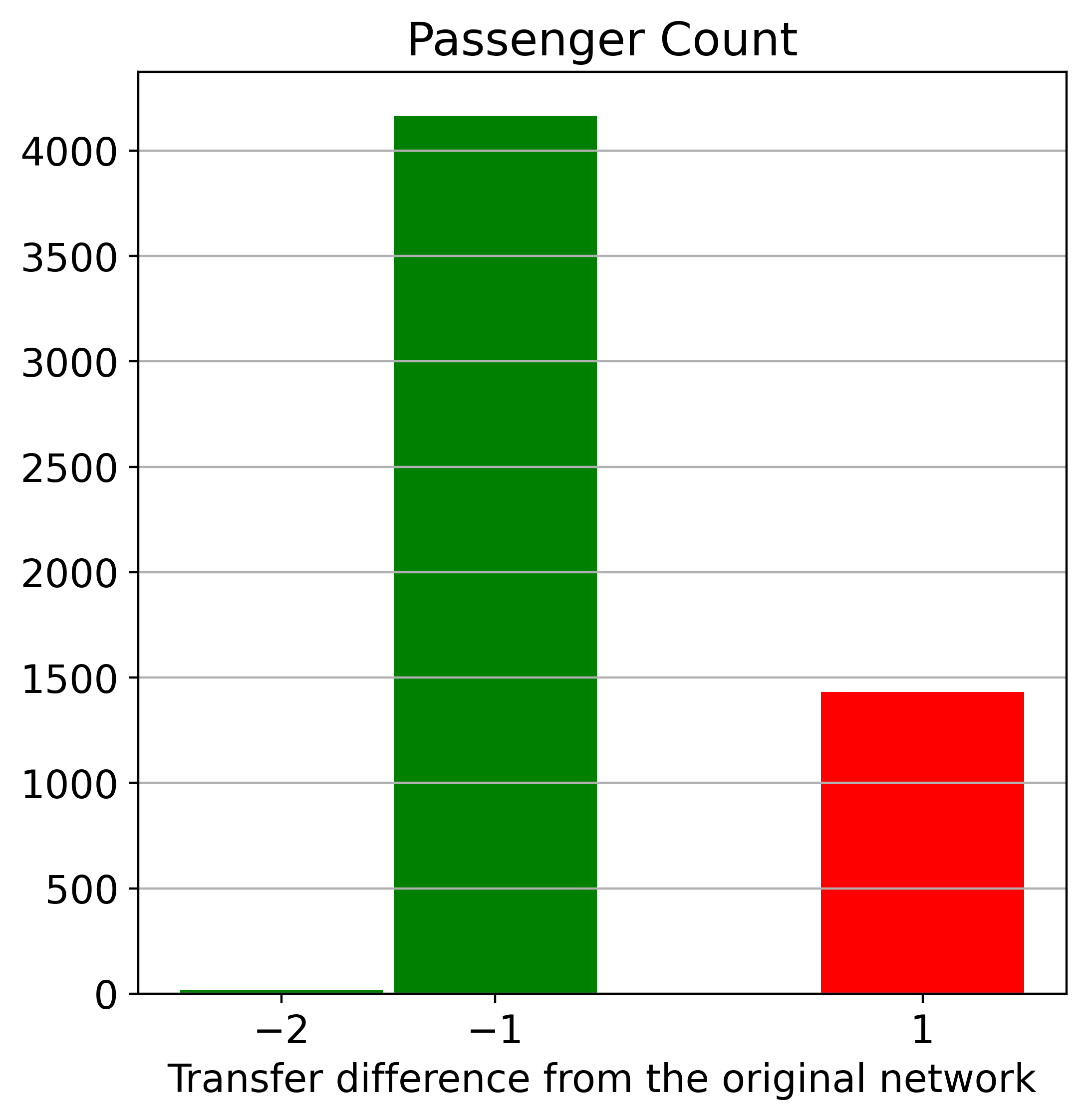}
    \vskip -0.1cm
    \caption{\small Differences in \textit{transfers} from the original network to the best network.}
    \label{fig:trasnfersdifference}
    \vskip -0.3cm
\end{figure}

The best network obtained through optimization has 200 routes, 109 less than the 
original network which translates in a difference of around $1300Km$ that no longer
have to be traveled by the buses. In Figure \ref{fig:traveltimedifference} 
we can see the differences in travel times imposed by the new network. Most of
the passengers has their traveling time unaffected (differences of up to $2.5$
are not considered) but the majority of trips in which there are significant changes
were positively affected. The network also enabled more direct trips as seen in
Figure \ref{fig:trasnfersdifference}. No passenger had their trips increased in more
than one stage and the passengers that had stages cut off their trips are more than
double of the ones that have to transfer one additional time now, some even cut 
two transfers off their trips. In these histograms, the central bar is omitted because
the total amount of passengers is $566K$, which would make the differences between the
smaller bars imperceptible. For the big majority of passengers, travel times and 
transfers remained the same. This is due to the big majority of traffic happening 
in the center of Lisbon in which most trips are already direct with the existing 
network. The new network is not able to satisfy as many trips as the original 
network but the difference is marginal. In general, the new network is able to
provide a similar service to the one provided in the original network, but does so
in a much more efficient way, using less routes.

\section{Conclusion}

In this work, we propose an end-to-end system to redesign a (multimodal) transport network using OD demand inferred from smartcard validations. 
The target system synergistically integrates the potentialities of multiple and single objective optimization approaches to address the core difficulty associated with the formulation of weighting schema. The TNDP problem is modeled as an evolutionary problem, where solutions in the Pareto front are firstly identified with a multi objective algorithm. A comparative rating of solutions drawn from the Pareto front is used to produce the weighting schema for single objective formulations through a (linear) regression model. 
In the context of our work, the targeted objectives are the unsatisfied demand, in-vehicle time, transfer needs, and the total length of the network as a proxy for the operator's costs and environmental footprint. We further consider safety capacity concerns and walking distances in transfers as parameterizable constraints.

Overall, the application of the proposed system over the Lisbon's public transport network showed moderate improvements in passenger-related objectives while massive improvements in operator-related objectives. 
According to the placed criteria, the best network in the Pareto Front approximation reduces the original network 
objective function in $23.1\%$ and the network built through single objective
optimization provides a $44.0\%$ reduction relative to the original network.
This last reduction translates to a $34.6\%$ reduction in total network length
(from 309 routes 200), a $5.1\%$ reduction in average transfers, a slight 
reduction of in-vehicle time and an increase of unsatisfied demand from  
$0.7\%$ to $1.3\%$. 
The transition from Multi Objective to Single Objective optimization 
proved effective. With a Single Objective optimization considering
the weights inferred through our rating process being able to produce
better networks than the best one in the Pareto Front after 60 iterations,
while a Single Objective optimization considering a uniform weighted sum
of all objectives normalized achieved the same only after 200 iterations.

This work shifts the focus from the subjective and difficult task of weighting conflicting objectives towards the objective task of rating networks in accordance with their ability to satisfy the placed objectives. In addition, the proposed system consistently combines state-of-the-art principles on network processing, origin-destination matrix inference, route generation, and evolutionary approaches to optimization. Grounded on automated fare collection data, the system can be used to dynamically adapt the transport network in the presence of more recent traffic data, being able to handle ongoing changes to the city traffic.

As lines of future work, we highlight the evaluation of the proposed methodology in other urban transport systems; 
a comprehensive assessment of the impact of placing objectives as bounded constraints in the problem formulation; 
and the establishment of principles to assist with the necessary incremental network reforms to guarantee the successful translation of the redesigned networks. 
\vskip 0.5cm
\noindent\textbf{Acknowledgments}. This work is supported by national funds through FCT under project ILU ({DSAIPA/DS/0111/2018}) and the INESC-ID pluriannual ({UIDB/50021/2020}).
\vskip 5cm

\bibliographystyle{ACM-Reference-Format}
\bibliography{cas-refs}


\begin{thebibliography}{24}


\ifx \showCODEN    \undefined \def \showCODEN     #1{\unskip}     \fi
\ifx \showDOI      \undefined \def \showDOI       #1{#1}\fi
\ifx \showISBNx    \undefined \def \showISBNx     #1{\unskip}     \fi
\ifx \showISBNxiii \undefined \def \showISBNxiii  #1{\unskip}     \fi
\ifx \showISSN     \undefined \def \showISSN      #1{\unskip}     \fi
\ifx \showLCCN     \undefined \def \showLCCN      #1{\unskip}     \fi
\ifx \shownote     \undefined \def \shownote      #1{#1}          \fi
\ifx \showarticletitle \undefined \def \showarticletitle #1{#1}   \fi
\ifx \showURL      \undefined \def \showURL       {\relax}        \fi
\providecommand\bibfield[2]{#2}
\providecommand\bibinfo[2]{#2}
\providecommand\natexlab[1]{#1}
\providecommand\showeprint[2][]{arXiv:#2}

\bibitem[\protect\citeauthoryear{Arbex and da~Cunha}{Arbex and
  da~Cunha}{2015}]%
        {arbex2015efficient}
\bibfield{author}{\bibinfo{person}{Renato~Oliveira Arbex} {and}
  \bibinfo{person}{Claudio~Barbieri da Cunha}.}
  \bibinfo{year}{2015}\natexlab{}.
\newblock \showarticletitle{Efficient transit network design and frequencies
  setting multi-objective optimization by alternating objective genetic
  algorithm}.
\newblock \bibinfo{journal}{\emph{Transportation Research Part B:
  Methodological}}  \bibinfo{volume}{81} (\bibinfo{year}{2015}),
  \bibinfo{pages}{355--376}.
\newblock


\bibitem[\protect\citeauthoryear{Baaj and Mahmassani}{Baaj and
  Mahmassani}{1991}]%
        {baaj1991ai}
\bibfield{author}{\bibinfo{person}{M~Hadi Baaj} {and} \bibinfo{person}{Hani~S
  Mahmassani}.} \bibinfo{year}{1991}\natexlab{}.
\newblock \showarticletitle{An AI-based approach for transit route system
  planning and design}.
\newblock \bibinfo{journal}{\emph{Journal of advanced transportation}}
  \bibinfo{volume}{25}, \bibinfo{number}{2} (\bibinfo{year}{1991}),
  \bibinfo{pages}{187--209}.
\newblock


\bibitem[\protect\citeauthoryear{Bielli, Caramia, and Carotenuto}{Bielli
  et~al\mbox{.}}{2002}]%
        {bielli2002genetic}
\bibfield{author}{\bibinfo{person}{Maurizio Bielli},
  \bibinfo{person}{Massimiliano Caramia}, {and} \bibinfo{person}{Pasquale
  Carotenuto}.} \bibinfo{year}{2002}\natexlab{}.
\newblock \showarticletitle{Genetic algorithms in bus network optimization}.
\newblock \bibinfo{journal}{\emph{Transportation Research Part C: Emerging
  Technologies}} \bibinfo{volume}{10}, \bibinfo{number}{1}
  (\bibinfo{year}{2002}), \bibinfo{pages}{19--34}.
\newblock


\bibitem[\protect\citeauthoryear{Brauers, Zavadskas, Peldschus, and
  Turskis}{Brauers et~al\mbox{.}}{2008}]%
        {brauers2008multi}
\bibfield{author}{\bibinfo{person}{Willem Karel~M Brauers},
  \bibinfo{person}{Edmundas~Kazimieras Zavadskas}, \bibinfo{person}{Friedel
  Peldschus}, {and} \bibinfo{person}{Zenonas Turskis}.}
  \bibinfo{year}{2008}\natexlab{}.
\newblock \showarticletitle{Multi-objective decision-making for road design}.
\newblock \bibinfo{journal}{\emph{Transport}} \bibinfo{volume}{23},
  \bibinfo{number}{3} (\bibinfo{year}{2008}), \bibinfo{pages}{183--193}.
\newblock


\bibitem[\protect\citeauthoryear{Cerqueira, Arsénio, and Henriques}{Cerqueira
  et~al\mbox{.}}{2021}]%
        {cerqueira2020estimators}
\bibfield{author}{\bibinfo{person}{Sofia Cerqueira}, \bibinfo{person}{Elisabete
  Arsénio}, {and} \bibinfo{person}{Rui Henriques}.}
  \bibinfo{year}{2021}\natexlab{}.
\newblock \showarticletitle{Inference of Dynamic Origin-Destination Matrices
  with Trip and Transfer Status from Individual Smart Card Data}.
\newblock \bibinfo{journal}{\emph{European Transport Conference (ETC), Milan,
  Italy}} (\bibinfo{year}{2021}).
\newblock


\bibitem[\protect\citeauthoryear{Chen, Kim, Lee, and Kim}{Chen
  et~al\mbox{.}}{2010}]%
        {chen2010stochastic}
\bibfield{author}{\bibinfo{person}{Anthony Chen}, \bibinfo{person}{Juyoung
  Kim}, \bibinfo{person}{Seungjae Lee}, {and} \bibinfo{person}{Youngchan Kim}.}
  \bibinfo{year}{2010}\natexlab{}.
\newblock \showarticletitle{Stochastic multi-objective models for network
  design problem}.
\newblock \bibinfo{journal}{\emph{Expert Systems with Applications}}
  \bibinfo{volume}{37}, \bibinfo{number}{2} (\bibinfo{year}{2010}),
  \bibinfo{pages}{1608--1619}.
\newblock


\bibitem[\protect\citeauthoryear{Chen, Subprasom, and Ji}{Chen
  et~al\mbox{.}}{2006}]%
        {chen2006simulation}
\bibfield{author}{\bibinfo{person}{Anthony Chen}, \bibinfo{person}{Kitti
  Subprasom}, {and} \bibinfo{person}{Zhaowang Ji}.}
  \bibinfo{year}{2006}\natexlab{}.
\newblock \showarticletitle{A simulation-based multi-objective genetic
  algorithm (SMOGA) procedure for BOT network design problem}.
\newblock \bibinfo{journal}{\emph{Optimization and Engineering}}
  \bibinfo{volume}{7}, \bibinfo{number}{3} (\bibinfo{year}{2006}),
  \bibinfo{pages}{225--247}.
\newblock


\bibitem[\protect\citeauthoryear{Chien, Yang, and Hou}{Chien
  et~al\mbox{.}}{2001}]%
        {chien2001genetic}
\bibfield{author}{\bibinfo{person}{Steven Chien}, \bibinfo{person}{Zhaowei
  Yang}, {and} \bibinfo{person}{Edwin Hou}.} \bibinfo{year}{2001}\natexlab{}.
\newblock \showarticletitle{Genetic algorithm approach for transit route
  planning and design}.
\newblock \bibinfo{journal}{\emph{Journal of transportation engineering}}
  \bibinfo{volume}{127}, \bibinfo{number}{3} (\bibinfo{year}{2001}),
  \bibinfo{pages}{200--207}.
\newblock


\bibitem[\protect\citeauthoryear{Deb, Pratap, Agarwal, and Meyarivan}{Deb
  et~al\mbox{.}}{2002}]%
        {deb2002fast}
\bibfield{author}{\bibinfo{person}{Kalyanmoy Deb}, \bibinfo{person}{Amrit
  Pratap}, \bibinfo{person}{Sameer Agarwal}, {and} \bibinfo{person}{TAMT
  Meyarivan}.} \bibinfo{year}{2002}\natexlab{}.
\newblock \showarticletitle{A fast and elitist multiobjective genetic
  algorithm: NSGA-II}.
\newblock \bibinfo{journal}{\emph{IEEE transactions on evolutionary
  computation}} \bibinfo{volume}{6}, \bibinfo{number}{2}
  (\bibinfo{year}{2002}), \bibinfo{pages}{182--197}.
\newblock


\bibitem[\protect\citeauthoryear{{Fan} and {Machemehl}}{{Fan} and
  {Machemehl}}{2006}]%
        {fan2006optimal}
\bibfield{author}{\bibinfo{person}{Wei {Fan}} {and} \bibinfo{person}{Randy~B.
  {Machemehl}}.} \bibinfo{year}{2006}\natexlab{}.
\newblock \showarticletitle{Optimal Transit Route Network Design Problem with
  Variable Transit Demand: Genetic Algorithm Approach}.
\newblock \bibinfo{journal}{\emph{Journal of Transportation Engineering-asce}}
  \bibinfo{volume}{132}, \bibinfo{number}{1} (\bibinfo{year}{2006}),
  \bibinfo{pages}{40--51}.
\newblock


\bibitem[\protect\citeauthoryear{{Farahani}, {Miandoabchi}, {Szeto}, and
  {Rashidi}}{{Farahani} et~al\mbox{.}}{2013}]%
        {farahani2013a}
\bibfield{author}{\bibinfo{person}{Reza~Zanjirani {Farahani}},
  \bibinfo{person}{Elnaz {Miandoabchi}}, \bibinfo{person}{W.Y. {Szeto}}, {and}
  \bibinfo{person}{Hannaneh {Rashidi}}.} \bibinfo{year}{2013}\natexlab{}.
\newblock \showarticletitle{A review of urban transportation network design
  problems}.
\newblock \bibinfo{journal}{\emph{European Journal of Operational Research}}
  \bibinfo{volume}{229}, \bibinfo{number}{2} (\bibinfo{year}{2013}),
  \bibinfo{pages}{281--302}.
\newblock


\bibitem[\protect\citeauthoryear{Inti and Kumar}{Inti and Kumar}{2021}]%
        {inti2021sustainable}
\bibfield{author}{\bibinfo{person}{Sundeep Inti} {and}
  \bibinfo{person}{Siddagangaiah~Anjan Kumar}.}
  \bibinfo{year}{2021}\natexlab{}.
\newblock \showarticletitle{Sustainable road design through multi-objective
  optimization: A case study in Northeast India}.
\newblock \bibinfo{journal}{\emph{Transportation research part D: transport and
  environment}}  \bibinfo{volume}{91} (\bibinfo{year}{2021}),
  \bibinfo{pages}{102594}.
\newblock


\bibitem[\protect\citeauthoryear{Jha, Jha, and Tiwari}{Jha
  et~al\mbox{.}}{2019}]%
        {jha2019multi}
\bibfield{author}{\bibinfo{person}{Shashi~Bhushan Jha},
  \bibinfo{person}{Jitendra~Kumar Jha}, {and} \bibinfo{person}{Manoj~Kumar
  Tiwari}.} \bibinfo{year}{2019}\natexlab{}.
\newblock \showarticletitle{A multi-objective meta-heuristic approach for
  transit network design and frequency setting problem in a bus transit
  system}.
\newblock \bibinfo{journal}{\emph{Computers \& Industrial Engineering}}
  \bibinfo{volume}{130} (\bibinfo{year}{2019}), \bibinfo{pages}{166--186}.
\newblock


\bibitem[\protect\citeauthoryear{Jia, Ma, and Hu}{Jia et~al\mbox{.}}{2019}]%
        {jia2019urban}
\bibfield{author}{\bibinfo{person}{Guo-Ling Jia}, \bibinfo{person}{Rong-Guo
  Ma}, {and} \bibinfo{person}{Zhi-Hua Hu}.} \bibinfo{year}{2019}\natexlab{}.
\newblock \showarticletitle{Urban transit network properties evaluation and
  optimization based on complex network theory}.
\newblock \bibinfo{journal}{\emph{Sustainability}} \bibinfo{volume}{11},
  \bibinfo{number}{7} (\bibinfo{year}{2019}), \bibinfo{pages}{2007}.
\newblock


\bibitem[\protect\citeauthoryear{Miandoabchi, Farahani, Dullaert, and
  Szeto}{Miandoabchi et~al\mbox{.}}{2012}]%
        {miandoabchi2012hybrid}
\bibfield{author}{\bibinfo{person}{Elnaz Miandoabchi},
  \bibinfo{person}{Reza~Zanjirani Farahani}, \bibinfo{person}{Wout Dullaert},
  {and} \bibinfo{person}{Wai~Yuen Szeto}.} \bibinfo{year}{2012}\natexlab{}.
\newblock \showarticletitle{Hybrid evolutionary metaheuristics for concurrent
  multi-objective design of urban road and public transit networks}.
\newblock \bibinfo{journal}{\emph{Networks and Spatial Economics}}
  \bibinfo{volume}{12}, \bibinfo{number}{3} (\bibinfo{year}{2012}),
  \bibinfo{pages}{441--480}.
\newblock


\bibitem[\protect\citeauthoryear{Newell}{Newell}{1979}]%
        {newell1979some}
\bibfield{author}{\bibinfo{person}{Gordon~F Newell}.}
  \bibinfo{year}{1979}\natexlab{}.
\newblock \showarticletitle{Some issues relating to the optimal design of bus
  routes}.
\newblock \bibinfo{journal}{\emph{Transportation Science}}
  \bibinfo{volume}{13}, \bibinfo{number}{1} (\bibinfo{year}{1979}),
  \bibinfo{pages}{20--35}.
\newblock


\bibitem[\protect\citeauthoryear{Pattnaik, Mohan, and Tom}{Pattnaik
  et~al\mbox{.}}{1998}]%
        {pattnaik1998urban}
\bibfield{author}{\bibinfo{person}{SB Pattnaik}, \bibinfo{person}{S Mohan},
  {and} \bibinfo{person}{VM Tom}.} \bibinfo{year}{1998}\natexlab{}.
\newblock \showarticletitle{Urban bus transit route network design using
  genetic algorithm}.
\newblock \bibinfo{journal}{\emph{Journal of transportation engineering}}
  \bibinfo{volume}{124}, \bibinfo{number}{4} (\bibinfo{year}{1998}),
  \bibinfo{pages}{368--375}.
\newblock


\bibitem[\protect\citeauthoryear{Sampson}{Sampson}{1976}]%
        {geneticalgorithms}
\bibfield{author}{\bibinfo{person}{Jeffrey~R Sampson}.}
  \bibinfo{year}{1976}\natexlab{}.
\newblock \bibinfo{title}{Adaptation in natural and artificial systems (John H.
  Holland)}.
\newblock
\newblock


\bibitem[\protect\citeauthoryear{Sohn}{Sohn}{2011}]%
        {sohn2011multi}
\bibfield{author}{\bibinfo{person}{Keemin Sohn}.}
  \bibinfo{year}{2011}\natexlab{}.
\newblock \showarticletitle{Multi-objective optimization of a road diet network
  design}.
\newblock \bibinfo{journal}{\emph{Transportation research part A: policy and
  practice}} \bibinfo{volume}{45}, \bibinfo{number}{6} (\bibinfo{year}{2011}),
  \bibinfo{pages}{499--511}.
\newblock


\bibitem[\protect\citeauthoryear{Vuurstaek, Cich, Knapen, Ectors, Bellemans,
  Janssens, et~al\mbox{.}}{Vuurstaek et~al\mbox{.}}{2018}]%
        {vuurstaek2018gtfs}
\bibfield{author}{\bibinfo{person}{Jan Vuurstaek}, \bibinfo{person}{Glenn
  Cich}, \bibinfo{person}{Luk Knapen}, \bibinfo{person}{Wim Ectors},
  \bibinfo{person}{Tom Bellemans}, \bibinfo{person}{Davy Janssens},
  {et~al\mbox{.}}} \bibinfo{year}{2018}\natexlab{}.
\newblock \showarticletitle{GTFS bus stop mapping to the OSM network}.
\newblock \bibinfo{journal}{\emph{Future Generation Computer Systems}}
  (\bibinfo{year}{2018}).
\newblock


\bibitem[\protect\citeauthoryear{Wan and Lo}{Wan and Lo}{2003}]%
        {wan2003mixed}
\bibfield{author}{\bibinfo{person}{Quentin~K Wan} {and} \bibinfo{person}{Hong~K
  Lo}.} \bibinfo{year}{2003}\natexlab{}.
\newblock \showarticletitle{A mixed integer formulation for multiple-route
  transit network design}.
\newblock \bibinfo{journal}{\emph{Journal of Mathematical Modelling and
  Algorithms}} \bibinfo{volume}{2}, \bibinfo{number}{4} (\bibinfo{year}{2003}),
  \bibinfo{pages}{299--308}.
\newblock


\bibitem[\protect\citeauthoryear{Wang, Ye, and Wang}{Wang
  et~al\mbox{.}}{2020}]%
        {wang2020multi}
\bibfield{author}{\bibinfo{person}{Chao Wang}, \bibinfo{person}{Zhirui Ye},
  {and} \bibinfo{person}{Wei Wang}.} \bibinfo{year}{2020}\natexlab{}.
\newblock \showarticletitle{A multi-objective optimization and hybrid heuristic
  approach for urban bus route network design}.
\newblock \bibinfo{journal}{\emph{IEEE Access}}  \bibinfo{volume}{8}
  (\bibinfo{year}{2020}), \bibinfo{pages}{12154--12167}.
\newblock


\bibitem[\protect\citeauthoryear{Yu, Yang, Cheng, and Liu}{Yu
  et~al\mbox{.}}{2005}]%
        {yu2005optimizing}
\bibfield{author}{\bibinfo{person}{Bin Yu}, \bibinfo{person}{Zhongzhen Yang},
  \bibinfo{person}{Chuntian Cheng}, {and} \bibinfo{person}{Chong Liu}.}
  \bibinfo{year}{2005}\natexlab{}.
\newblock \showarticletitle{Optimizing bus transit network with parallel ant
  colony algorithm}. In \bibinfo{booktitle}{\emph{Proceedings of the Eastern
  Asia Society for Transportation Studies}}, Vol.~\bibinfo{volume}{5}.
  Citeseer, \bibinfo{pages}{374--389}.
\newblock


\bibitem[\protect\citeauthoryear{Zhang, Pu, Schonfeld, Song, Li, Wang, Peng,
  and Hu}{Zhang et~al\mbox{.}}{2020}]%
        {zhang2020multi}
\bibfield{author}{\bibinfo{person}{Hong Zhang}, \bibinfo{person}{Hao Pu},
  \bibinfo{person}{Paul Schonfeld}, \bibinfo{person}{Taoran Song},
  \bibinfo{person}{Wei Li}, \bibinfo{person}{Jie Wang},
  \bibinfo{person}{Xianbao Peng}, {and} \bibinfo{person}{Jianping Hu}.}
  \bibinfo{year}{2020}\natexlab{}.
\newblock \showarticletitle{Multi-objective railway alignment optimization
  considering costs and environmental impacts}.
\newblock \bibinfo{journal}{\emph{Applied Soft Computing}}
  \bibinfo{volume}{89} (\bibinfo{year}{2020}), \bibinfo{pages}{106105}.
\newblock


\end{thebibliography}





\end{document}